\documentclass{article}

\usepackage{arxiv}

\usepackage[utf8]{inputenc} 
\usepackage[T1]{fontenc}    
\usepackage{hyperref}       
\usepackage{url}            
\usepackage{booktabs}       
\usepackage{amsfonts}       
\usepackage{nicefrac}       
\usepackage{microtype}      
\usepackage{lipsum}
\usepackage{graphicx}
\graphicspath{ {./images/} }

\usepackage{amsmath}
\usepackage{subfig}
\usepackage{algorithm}
\usepackage{algpseudocode}
\usepackage{array}
\newcolumntype{L}{>{\centering\arraybackslash}m{1.5cm}}
\newcolumntype{P}[1]{>{\centering\arraybackslash}p{#1}}

\newcommand\blfootnote[1]{%
  \begingroup
  \renewcommand\thefootnote{}\footnote{#1}%
  \addtocounter{footnote}{-1}%
  \endgroup
}

\title{Learning to Rank Intents in Voice Assistants}

\author{
 Raviteja Anantha \\
  Apple Inc.\\
  \texttt{raviteja\_anantha@apple.com} \\
   \And
 Srinivas Chappidi \\
  Apple Inc.\\
  \texttt{vasuc@apple.com} \\
  \And
 William Dawoodi \\
  Apple Inc.\\
  \texttt{dawoodi@apple.com} \\
}

\begin{document}
\maketitle
\begin{abstract}
Voice Assistants aim to fulfill user requests by choosing the best intent from multiple options generated by its Automated Speech Recognition and Natural Language Understanding sub-systems. However, voice assistants do not always produce the expected results. This can happen because voice assistants choose from ambiguous intents --- user-specific or domain-specific contextual information reduces the ambiguity of the user request. Additionally the user information-state can be leveraged to understand how relevant/executable a specific intent is for a user request. In this work, we propose a novel Energy-based model for the intent ranking task, where we learn an affinity metric and model the trade-off between extracted meaning from speech utterances and relevance/executability aspects of the intent. Furthermore we present a Multisource Denoising Autoencoder based pretraining that is capable of learning fused representations of data from multiple sources. We empirically show our approach outperforms existing state of the art methods by reducing the error-rate by 3.8\%, which in turn reduces ambiguity and eliminates undesired dead-ends leading to better user experience. Finally, we evaluate the robustness of our algorithm on the intent ranking task and show our algorithm improves the robustness by 33.3\%. \blfootnote{This work has been accepted at IWSDS 2020 Conference.}
\end{abstract}


\section{Introduction}
\label{intro}
\begin{figure*}
 \centering
  \includegraphics[scale=0.16]{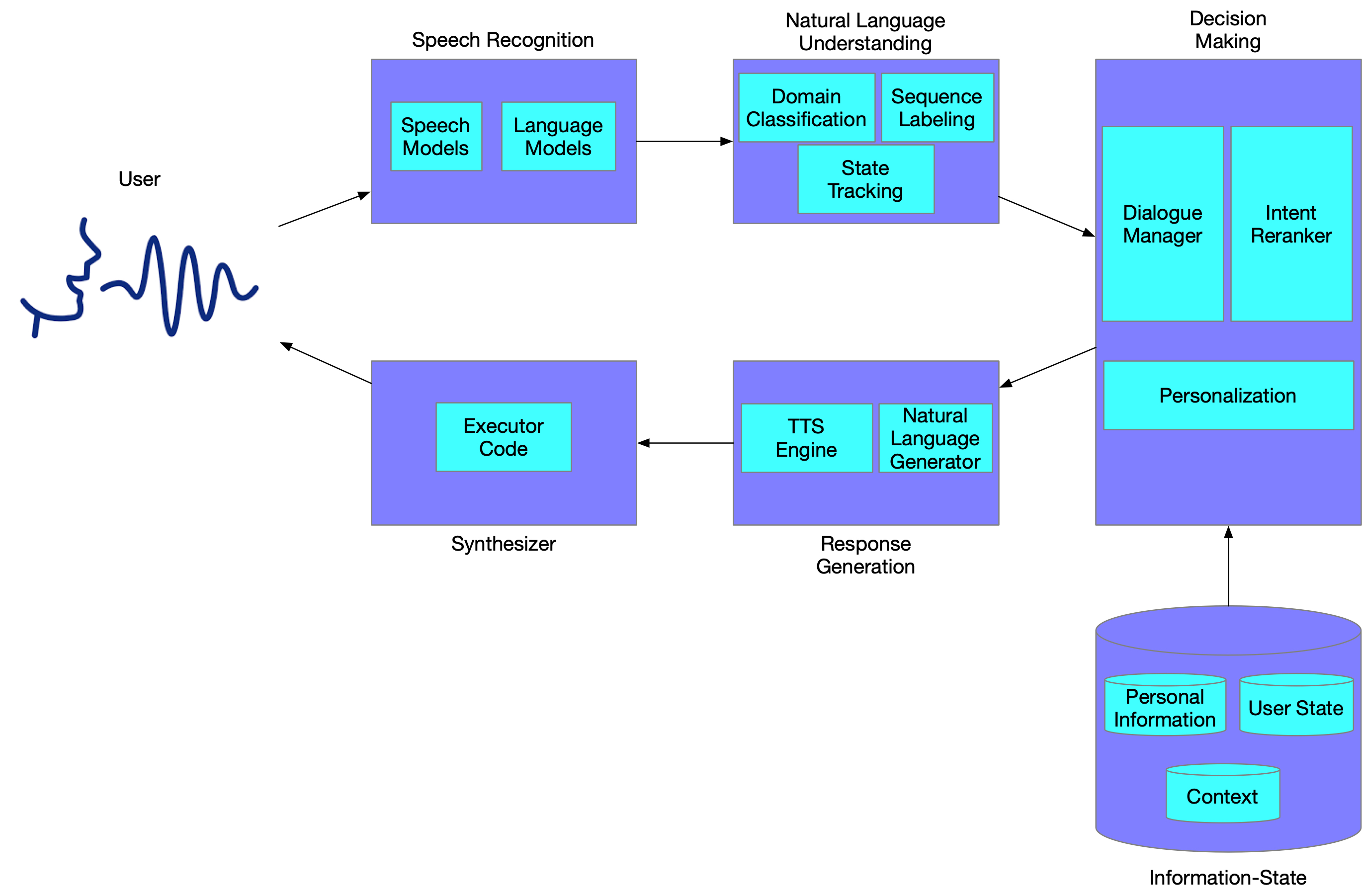}
  \caption{Components of a Voice Assistant.}
  \label{fig:components_voice_assistants}
\end{figure*}

A variety of tasks use Voice Assistants (VA) as their main user interface. VAs must overcome complex problems and hence they typically are formed of a number of components: one that transcribes the user speech (Automated Speech Recognition - ASR), one that understands the transcribed utterances (Natural Language Understanding - NLU), one that makes decisions (Decision Making - DM \cite{blaise:13}), and one that produces the output speech (TTS). Many VAs have a pipeline structure similar to that in Figure \ref{fig:components_voice_assistants}.

Our work is mainly focused on the DM sub-system and our primary contributions are: 1) proposing to decouple language understanding from information-state and modeling an affinity metric between them; 2) the identification of Multisource Denoising Autoencoder based pretraining and its application to learn robust fused representations; 3) quantifying robustness; 4) the introduction of a novel ranking algorithm using Energy-based models (EBMs). In this work, we limit our scope to non-conversational utterances, \textit{i.e.}, utterances without followups containing anaphoric references and leave that for future work. We evaluate our approach on an internal dataset. Since our algorithm is primarily focused on leveraging inherent characteristics that are unique to large-scale real-world VAs, the exact algorithm may not be directly applicable to open-source \textit{Learning to Rank} (LTR) datasets. But we hope our findings will encourage application and exploration of EBMs applied to LTR in both real-world VAs and other LTR settings.

The remainder of the paper is organized as follows: Section \ref{task} discusses the task description while Section \ref{work} covers the related work. Section \ref{algorithm} then describes the ranking algorithm, and Section \ref{exps} discusses the evaluation metrics, datasets, training procedure, and results. 

\section{Task Description}
\label{task}
The ultimate goal of a VA is to understand user intent. The exact meaning of the words is often not enough to choose the best intent. In Figure \ref{fig:components_voice_assistants}, we show the use of information-state, and we classify it into three categories. All private-sensitive information stays on the user's device.

\textbf{Personal Information:}
\textit{e.g.} user-location, app subscriptions, browsing history, device-type etc.

\textbf{User State:}
Information about the user's state at the time a query is made. (\textit{e.g.} user is driving, etc.)

\textbf{Context:}
Dialog context of what the user said in previous queries in the same conversation or task (\textit{e.g.} song requests).

\begin{figure*}
 \centering
  \includegraphics[scale=0.15]{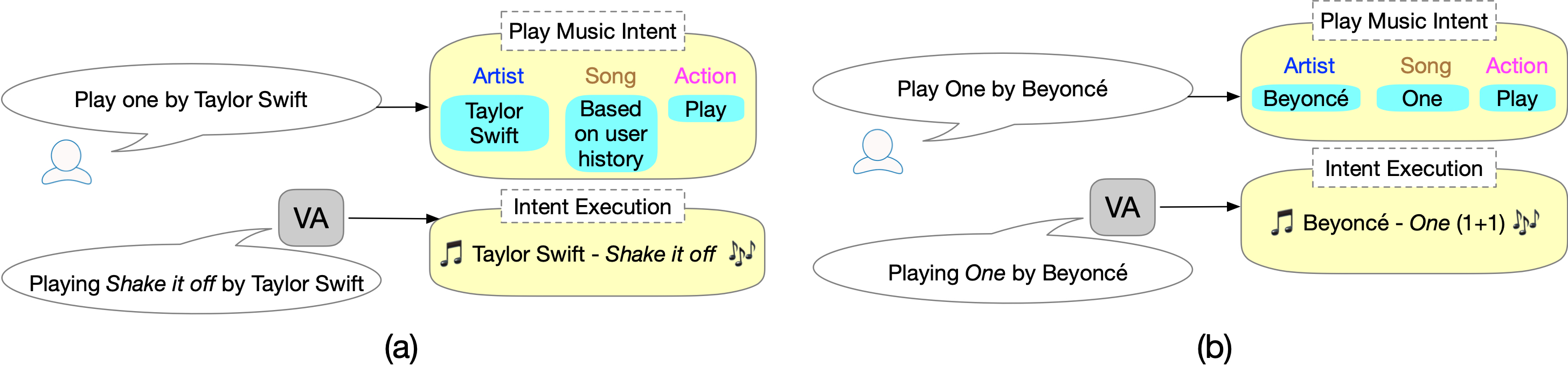}
  \caption{Examples of user requests with same semantics but with different intents. (a) shows a user request to play a song from an artist, (b) shows a user request to play a specific song from an artist.}
  \label{fig:one}
\end{figure*}

To illustrate how semantically similar user requests can have different user intents consider the examples in Figure \ref{fig:one}. In Figure \ref{fig:one}a the user meant to play some song from a specific artist. However in Figure \ref{fig:one}b, although playing some song from the requested artist is also reasonable, knowing that there is a song named ``One" from the artist leads to better intent selection, as shown. 

\begin{figure*}
 \centering
   \includegraphics[scale=0.15]{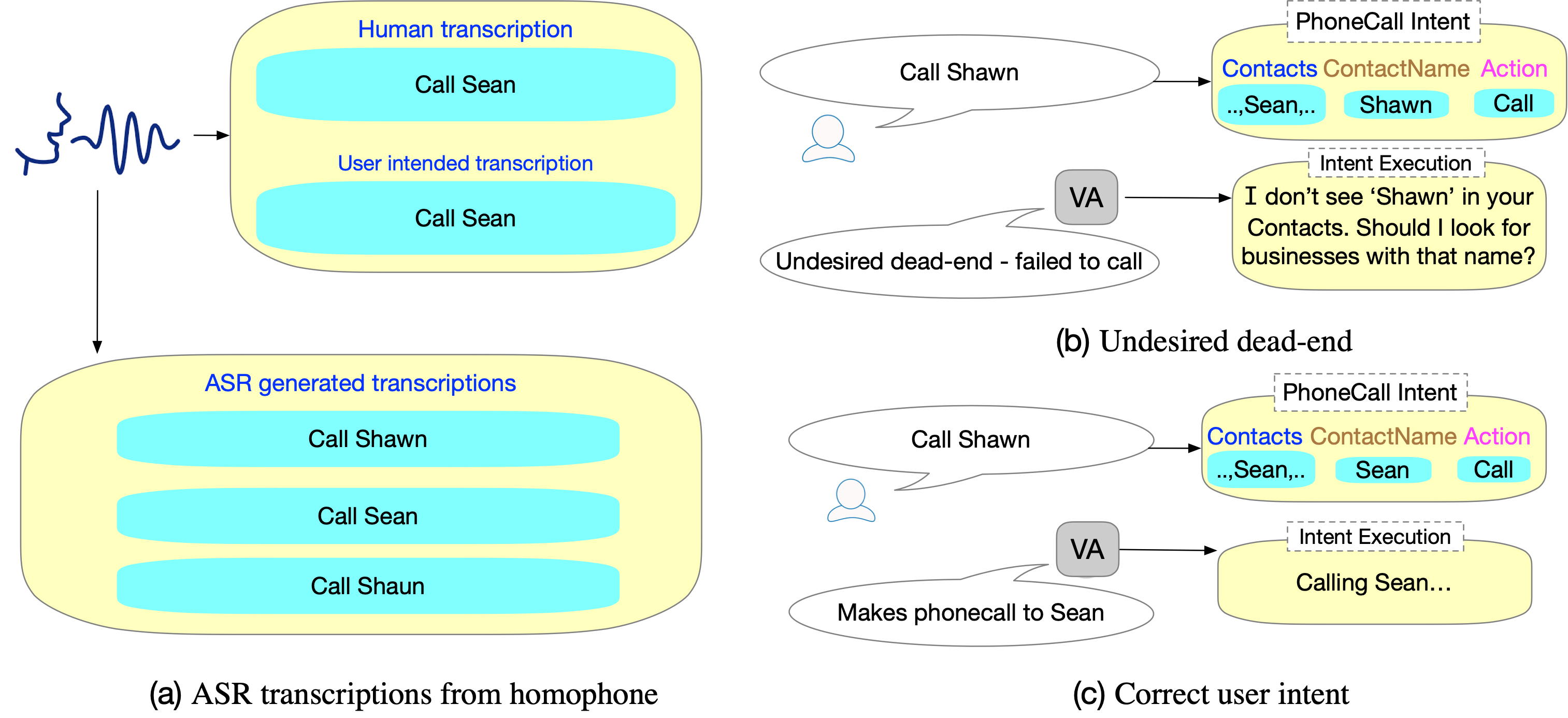}
   \caption{An example of an undesired dead-end. (a) shows a case where user intended transcription is not possible to predict unless the voice assistant has the contact information. (b) shows how lack of contact information leads to a sub-optimal intent execution although there is a better intent shown in (c).}
  \label{fig:uduc}
\end{figure*}

Ambiguity can still remain even if a sub-system correctly decodes user input. For example consider Figure \ref{fig:uduc}: it is not possible to predict the user intended transcription unless we know there is a contact with that name due to the homophone. Figure \ref{fig:uduc}b is an example where a suboptimal intent was executed although there was a  better intent as shown in Figure \ref{fig:uduc}c. We term this scenario \textit{undesired dead-end} since the user's intended task hit a dead-end.

The use of information-state is crucial to select the right response, which is also shown empirically in Section \ref{error-rate-sec}. We aim to reduce ambiguity (both ASR and NLU), and undesired dead-ends to improve the selection of the right intent by ranking alternative intents. ASR signals are comprised of speech and language features that generate speech lattices, model scores, text, etc. NLU signals are comprised of domain classification features such as domain categories, domain scores, sequence labels of the user request transcription, etc. An intent is a combination of ASR and NLU signals. We refer to these signals as \textit{understanding signals} decoded by ASR and NLU sub-systems. Every intent is encoded into a vector space and this process is described in Section \ref{mdae}. Our task is to produce a ranked list of intents using information-state in addition to understanding signals to choose the best response.

\section{Related Work}
\label{work}
While our work falls within the broad literature of LTR, we position it in the context of information-state based reranking, unsupervised pretraining, zero-shot learning, and EBMs applied to the DM sub-system of a Voice Assistant.

\textbf{Information-state based reranking:}
Reranking approaches have been used in VAs to rerank intents to improve accuracy. Response category classification can be improved by reranking \textit{k}-best transcriptions from multiple ASR engines \cite{morbini:12}. ASR accuracy can be improved by reranking multiple ASR candidates by using their syntactic properties in Human-Computer Interaction \cite{roberto:13}. Reranking domain hypotheses is shown to improve domain classification accuracy over just using domain classifiers without reranking \cite{ruhi:18, ruhi:1}.

All of the above approaches only focus on ASR candidates or domain hypotheses, which are strongly biased towards the semantics of the user request. Although \cite{ruhi:18} exploits user preferences along with NLU interpretation, they treat both of them as a single entity (hypothesis). In our work, we explicitly learn an affinity metric between information-state and predicted meaning from the transcribed utterance to choose the appropriate response.  

\textbf{Unsupervised pretraining:}
DM input consists of multiple diverse sources. For example, speech lattices, textual information, scores from ASR and NLU models, and unstructured contextual information, to name a few. Each data type has distinct characteristics, and learning representations across data types that capture the meaning of the user request is important. One approach is to use a deep boltzmann machine for learning a generative model to encode such multisource features \cite{russ:12}. Few approaches learn initial representations from unlabeled data through pretraining \cite{roberto:13, ruhi:1}. Encoding can also be learned by optimizing a neural network classifier weights by minimizing the combined loss of an autoencoder and a classifier \cite{ranza:08}. Both pretraining and classification can be jointly learned from labeled and unlabeled data, labeled data loss is used to obtain pseudo-labels, and pretraining is done using the pseudo-loss \cite{lee:13}. Pretraining for intial representations can also be realized by using a CNN2CRF architecture for slot tagging using labeled data, and learning dependencies both within and between latent clusters of unseen words \cite{ruhi:08}.

Although these previous works address few aspects of the multisource data problem, none of them address the robustness of the learned representations. Since DM consumes the outputs of many sub-systems that may change their distributional properties, for instance through retraining, some degree of robustness is desired to not drastically affect the response selection.

To address both distinct data characteristics and robustness, we propose using a Denoising Autoencoder (DAE) \cite{vincent:08} with a hierarchical topology that uses separate encoders for each data type. The average reconstruction loss contains both a separate term to minimize the error for each encoder, and the fused representations. This provides an unsupervised method for learning meaningful underlying fused representations of the multisource input.

\begin{figure*}
 \centering
  \includegraphics[scale=0.15]{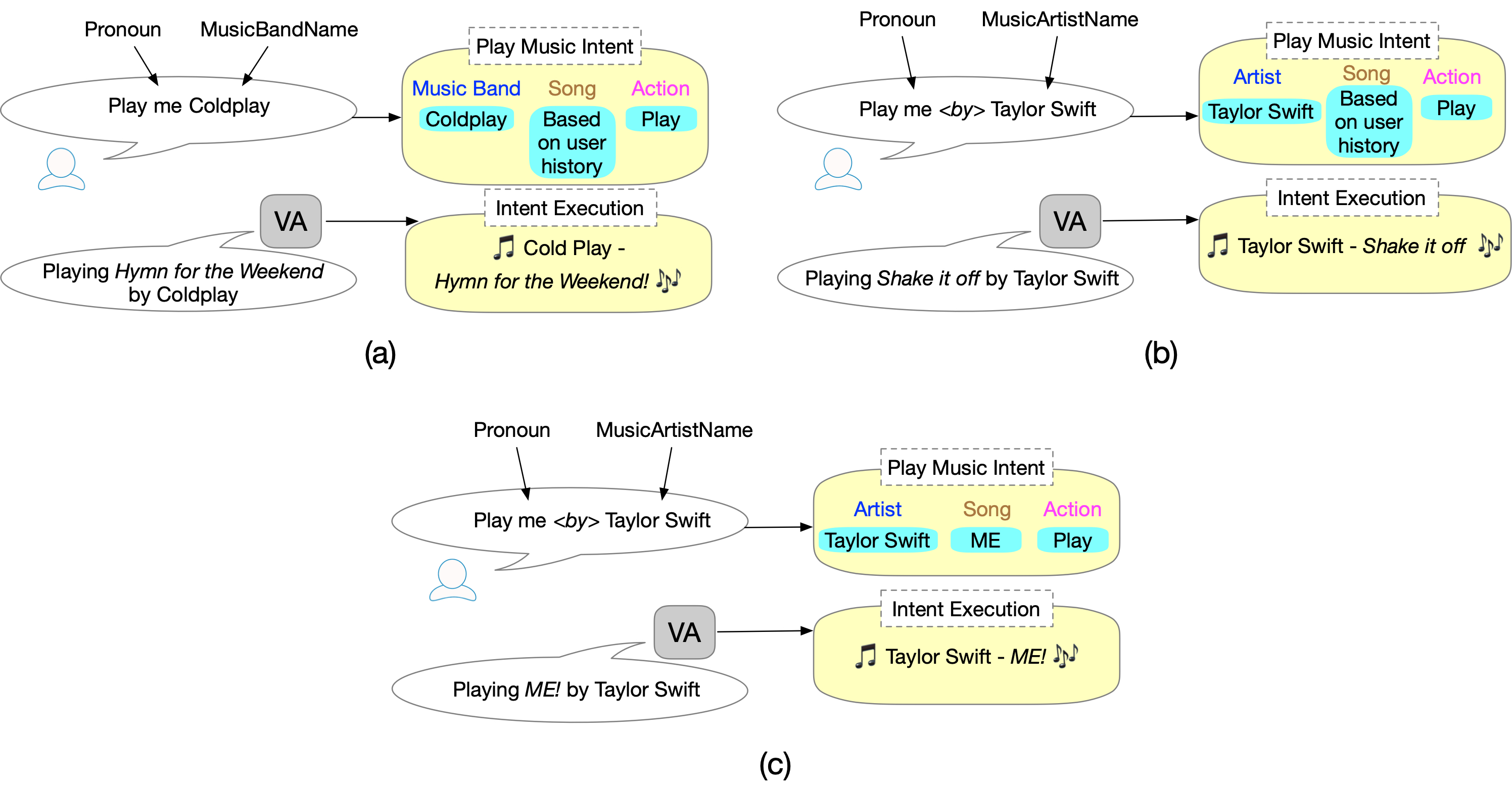}
  \caption{(a) Shows a scenario where NLU correctly predicts the intent given correct ASR transcription. (b) Shows a scenario where NLU fails to predict the right intent due to incorrect ASR transcription (missing the word ``by") caused by external noise. (c) Shows a scenario where NLU fails to predict the right intent, but DM helps in identifying the correct intent using domain-specific information.}
  \label{fig:me}
\end{figure*}

\textbf{Zero-shot learning:}
The ability of DM to predict and select unseen intents is important. User requests can consist of word sequences that NLU might not be able to accurately tag by relying only on language features. To illustrate consider the examples in Figure \ref{fig:me}. The user request in Figure \ref{fig:me}a is tagged correctly, and the NLU sub-system predicts the right user intent of playing a song from the correct artist. Figure \ref{fig:me}b showcases a scenario where due to external noise the user intended transcription of ``Play ME by Taylor Swift" was mistranscribed by the ASR sub-system as ``Play me Taylor Swift", and this ASR error propagated to NLU leading to tag \textit{ME} as a pronoun instead of \textit{MusicTitle}. With DM, as shown in Figure \ref{fig:me}c, we leverage domain-specific information and decode the right transcription and intent (playing ME song) from the affinity metric, although this input combination was never seen before by the model.

One approach is to use a convolutional deep structured semantic model (CDSSM), which performs zero-shot learning by jointly learning the representations for user intents and associated utterances \cite{dilek:1}. This approach is not scalable since such queries can have numerous variations, and they follow no semantic pattern. We propose to complement NLU features with domain-specific information to decode the right intent in addition to shared semantic signals.

\textbf{EBM for DM:}
Traditional approaches to LTR use discriminative methods. Our approach learns an affinity metric that captures dependencies and correlations between semantics and information-state of the user request. We accomplish this learning by associating a scalar energy (a measure of compatibility) to each configuration of the model parameters. This learning framework is known as \textit{energy-based learning} and is used in various computer vision applications, such as signature verification \cite{cowan:93}, face verification \cite{sumit:05}, and one-shot image recognition \cite{koch:15}. We apply EBM for LTR (and DM in voice assistants) for the first time. We propose a novel energy-based learning ranking loss function.

\section{EnergyRank algorithm}
\label{algorithm}
EBMs assign unnormalized energy to all possible configurations of the variables \cite{yann:05,hinton:03}. The advantage of EBMs over traditional probabilistic models, especially generative models, is that there is no need for estimating normalized probability distributions over the input space. This is efficient since we can avoid computing partition functions. Our algorithm consists of two phases --- pretraining and learning the ranking function, which are described in Sections \ref{mdae} and \ref{model} respectively.

\begin{figure*}
 \centering
  \includegraphics[scale=0.15]{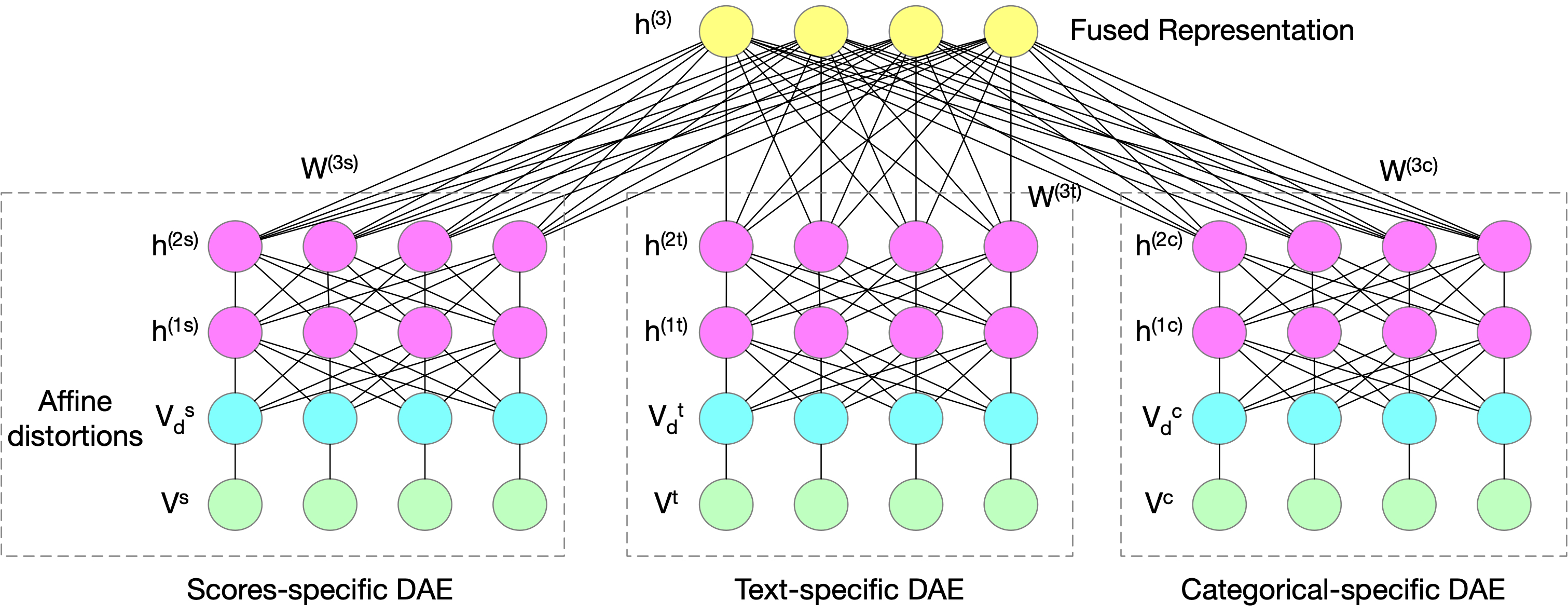}
  \caption{Encoder architecture of Multisource DAE that models the joint distribution over scores, text, and categorical features. \textit{Light green} layer, $V^{*}$, represents the original input; \textit{light magenta} layer, $V^{*}_{d}$, depicts the affine transformations; two layers of \textit{dark magenta}, $h^{1*}$ and  $h^{2*}$, represents source-specific latent representation learning; finally, \textit{light yellow} layer, $h^{(3)}$, represents the fused representation.}
  \label{fig:dae}
\end{figure*}

\subsection{Multisource Denoising Autoencoder}
\label{mdae}
Since our model consumes input from multiple sub-systems, two aspects are important: robustness of features and efficient encoding of multisource input. The concept of DAE \cite{vincent:08} is to be robust to variations of the input. We have three data types in the input: model scores that are produced by other sub-systems, text generated by ASR and Language Models (LMs), categorical features generated by NLU models like sequence labels, verbs etc. Let $V ^ {s}$ denote a multi-hot vector, which is a concatenation of 11 $ {\rm I\!R} ^ {11}$ one-hot vectors, where each contains binned real-valued model scores. Let $V ^ {t}$ represent the associated text input (padded or trimmed to a maximum of 20 words), which is a concatenation of 20 word-vectors. Each word-vector $v^{t}_{i} \in {\rm I\!R} ^ {50}$ is a multi-hot vector of $i^{th}$ word. Similarly let $V ^ {c}$ represent associated sequence-labels of those 20 words, which is a concatenation of 20 sequence-label vectors. Each $i^{th}$ sequence-label vector $v^{c}_{i} \in {\rm I\!R} ^ {50}$ is a multi-hot vector. For example consider the utterance ``Call Ravi'', the corresponding sequence-labels might be [\textit{phoneCallVerb}, \textit{contactName}].  

We start by modeling each data type by adding affine distortions followed by a separate two-layer projection of the encoder, as shown in Figure \ref{fig:dae}. This gives separate encodings for each data type. Let $dae_{*}$ represent an encoding function, $W^{*}_{enc}$ is the respective weight matrix and $P(noise)$ a uniform noise distribution. The encodings are  given by:
\begin{equation} \label{eq:1}
V^{s}_{d}, V^{t}_{d}, V^{c}_{d} = \textit{affine\_transform($(V^{s},V^{t},V^{c})$; $P(noise)$)}.
\end{equation}

Let us denote source-specific hidden representations of real-valued, text and categorical features by $h^{s}, h^{t}, h^{c}$ derived from encoder models with respective parameters $W^{s}_{enc}, W^{t}_{enc}, W^{c}_{enc}$. These latent representations are given by:
\begin{equation} \label{eq:2}
h^{*} = \textit{$dae_*$($V^{*}_{d}; W^{*}_{enc}$)},
\end{equation}
and the fused representation is obtained by:
\begin{equation} \label{eq:7}
h = \textit{$dae$($(h^{s},h^{t},h^{c}); W_{enc}$)}.
\end{equation}

Let $idae_{*}$ represent the decoding function, and $W^{*}_{dec}$ denote the respective weight matrix. The hidden-state reconstructions are given by:
\begin{equation} \label{eq:8}
h^{s'},h^{t'},h^{c'} = \textit{$idae$($h; (W^{s'}_{dec},W^{t'}_{dec},W^{c'}_{dec})$)}.
\end{equation}

The original denoised input reconstructions are given by:
\begin{equation} \label{eq:11}
V^{*'} = \textit{$idae_*$($h^{*'}; W^{*}_{dec}$)}.
\end{equation}

We learn the parameters of the Multisource DAE jointly by minimizing the average reconstruction error captured by \textit{categorical cross entropy} (CCE) of both the hidden state and the original denoised input decodings captured by the terms of the loss function. We denote the CCE loss as $L_{CCE}$. 
\begin{equation} \label{eq:14}
L^{h} = L_{CCE}(h^{*},h^{*'}),
\end{equation}
\begin{equation} \label{eq:15}
L^{V} = L_{CCE}(V^{*},V^{*'}),
\end{equation}
\begin{equation} \label{eq:16}
\textit{$W^{*}_{enc},W_{enc},W^{*}_{dec}$} = \underset{W^{*}_{enc}, W^{*}_{dec}}{\arg\min} \frac{1}{m}  \displaystyle\sum_{i=1}^{m} (L^{h}_{i} + L^{V}_{i}).
\end{equation}

\subsection{Model Description}
\label{model}
The ranking function is learned by finding the parameters $W$ that optimize the suitably designed ranking loss function evaluated over a validation set. Directly optimizing the loss averaged over an epoch generally leads to unstable EBM training and would be unlikely to converge \cite{sumit:05}. Therefore, we add a \textit{scoring layer} after the energy is computed and impose loss function forms to implicitly ensure energy is large for intent with bad rank and low otherwise. Details of the energy computation and the loss function forms are given in Sections \ref{ef} and \ref{efl} respectively. 

\begin{figure*}
 \centering
   \includegraphics[scale=0.14]{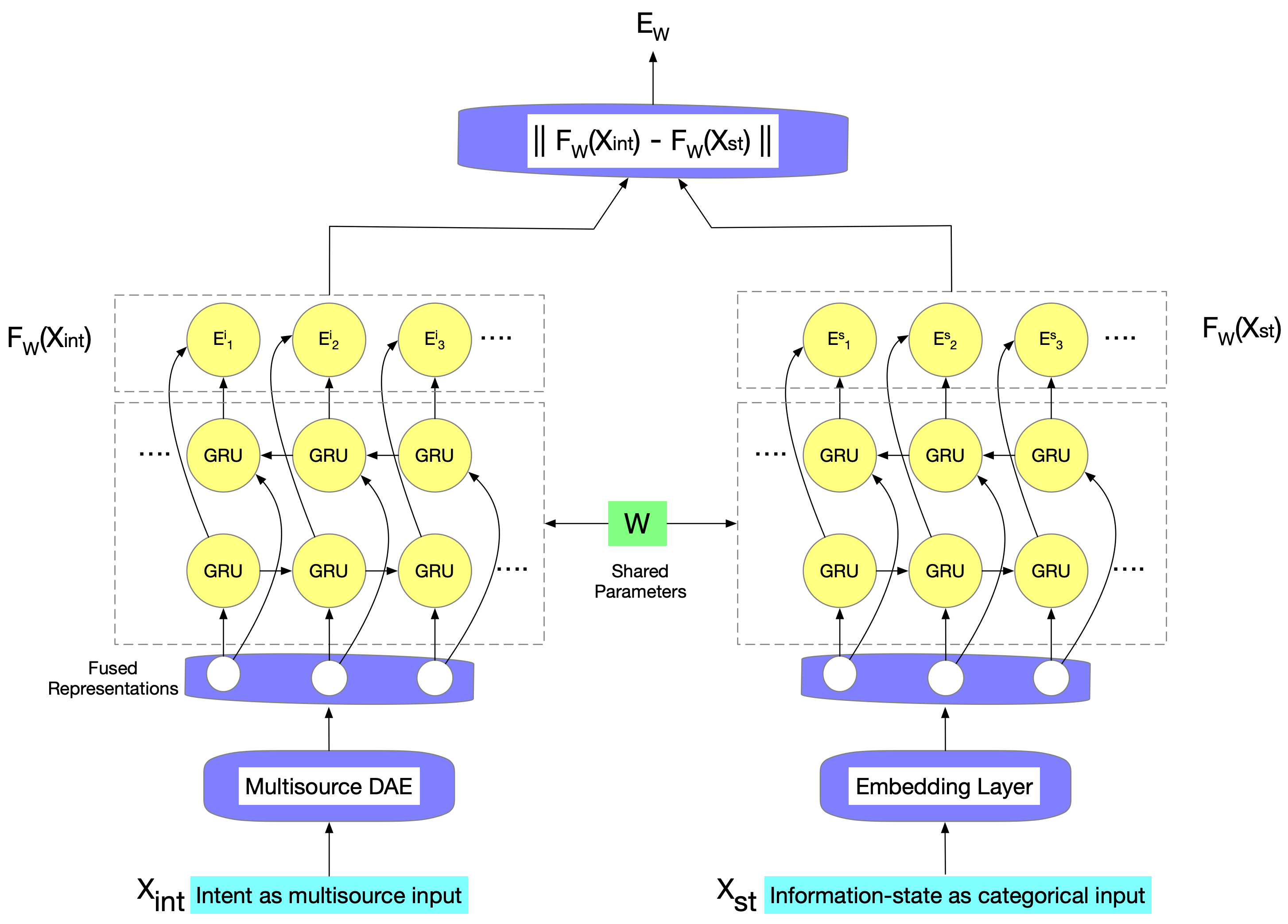}
   \caption{EBM with Siamese Architecture.}
  \label{fig:er}
\end{figure*}

\subsubsection{Energy Function of EBM}
\label{ef}
The architecture of our Ranker is shown in Figure \ref{fig:er}. Our ranker consists of two identical Bidirectional RNN networks, where one network accepts the fused representation, and the other accepts the information-state. Learning the affinity metric is realized by training these twin networks with shared weights. This type of architecture is called a Siamese Network \cite{cowan:93}. The major difference between our work and previous works on siamese networks is that we present the same data-point to the twin networks but categorized as two inputs based on if it is information-state or not. All previous works use two distinct data-points to compute energy. In other words, we compute intra-energy and previous works focused on inter-energy. We used GRU \cite{cho:14} for the RNN since it often has the same capacity as an LSTM \cite{hochreiter:97}, but with fewer parameters to train.

To simplify let $X_{int}$ and $X_{st}$ denote an intent's extracted meaning ($V^{s}, V^{t}, V^{c}$) and its associated information-state respectively. Both the inputs are transformed through Multisource DAE and Embeddings Layer respectively to have the same dimensions ${\rm I\!R} ^ {500}$. Let $W$ be the shared parameter matrix that is subject to learning, and let $F_{W}(X_{int})$ and $F_{W}(X_{st})$ be the two points in the metric space that are generated by mapping $X_{int}$ and $X_{st}$. The parameter matrix $W$ is shared even if the data sources of $X_{int}$ and $X_{st}$ are different since they are related to the same request and the model must learn the affinity between them. We compute the distance between $F_{W}(X_{int})$ and $F_{W}(X_{st})$ using the L1 norm, then the energy function that measures compatibility between $X_{int}$ and $X_{st}$ is defined as:

\begin{equation} \label{eq:17}
E_{W}(X_{int},X_{st}) = \Vert F_{W}(X_{int}) - F_{W}(X_{st}) \Vert.
\end{equation}

\subsubsection{Energy-based Ranking Loss Function}
\label{efl}
Traditional ranking loss functions construct the loss using some form of entropy in a pointwise, pairwise or listwise paradigm. Parameter updates are performed using either \textit{gradients} \cite{burges:05} or \textit{Lambdas} $\lambda$ \cite{burges:10, burges:06}. We use gradient based methods to update parameters. Let $x_{1}$ and $x_{2}$ be two intents from same user request. The prediction score of the ranker is obtained by $p = \sigma(E_{W})$, for convenience we denote $p$ associated with $x_1$ as $p(x_1)$ and $f(.)$ as the learned model function. We construct the loss as a sequence of weighted energy scores. Pairwise loss is constructed as:
\begin{equation} \label{eq:18}
L(f(.), x) = \sum_{i=1}^{n-1} \sum_{j=i+1}^{n} \phi (p(x_{i}), p(x_{j})),
\end{equation}
where $\phi$ is a hyperparameter that can be one of logistic function ($\phi(z) = log(1 + \exp^{-z})$), hinge function ($\phi(z) = (1-z)_{+}$), exponential function ($\phi(z) = \exp ^ {-z}$), with $z = p(x_{i}) - p(x_{j})$.

Listwise losses are constructed as:
\begin{equation} \label{eq:19}
L(p(.), x, y) = \sum_{i=1}^{n-1} (-p(x_{y(s)}) + ln(\sum_{j=i}^{n} \exp (p(x_{y(i)}))),
\end{equation}
where $y$ is a randomly selected permutation from the list of all possible intents that retains relevance to the user-request.

\section{Experiments and Results}
\label{exps}
\subsection{Evaluation Metrics}
We evaluated EnergyRank using two metrics. 
\begin{itemize}
\item \textbf{Error Rate} : The fraction of user requests where the intent selection was incorrect.

\item \textbf{Relative Entropy} : We employ \textit{Relative Entropy}, given in Equation \ref{eq:20}, to quantify the distance between input score distributions \textit{p} and \textit{q}. Relative entropy serves as a measure for the robustness of the model to upstream sub-system changes. We used \textit{whitening} to eliminate unbounded values, and 10E-9 as a dampening factor to give a bounded metric. A value of 0.0 indicates identical distributions, while 1.0 are maximally dissimilar.

\end{itemize}

\begin{equation}  \label{eq:20}
  rel\_entr(p,q) =
    \begin{cases}
      p \log{(p/q)} & p > 0, q > 0 \\
      0 & p=0, q \geq 0\\
      \infty & otherwise.
    \end{cases}       
\end{equation}

\subsection{Datasets}
\subsubsection{Labeled Dataset}
The labeled dataset is used to measure the error rate. This dataset contains 24,000 user requests comprised of seven domains: music, movies, app-launch, phone-call, and three knowledge-related domains. The ranking labels are produced by human annotators by taking non-private information-state into account. The dataset is divided into 12,000 user requests for training, 4,000 for validation and 8,000 for the test-set. The average number of predicted intents per user request is 9 with a maximum of 43. The extracted meaning of the request is represented by features from ASR and NLU sub-systems, information-state is represented by 114 categorical attributes. The error rate with just selecting the top hypothesis is 41\%.

\subsubsection{Unlabeled Dataset}
The unlabeled dataset consists of two unlabeled sub-datasets sampled from two different input distributions. Each sub-dataset consists of 80,000 user requests. The data here are not annotated since we are interested in a metric that only needs the scores of the model's best intent.

\subsection{Training Procedure}
We trained EBM using both pairwise and listwise loss functions given in Eq-\ref{eq:18} and Eq-\ref{eq:19} respectively. The objective is combined with backpropagation, where the gradient is additive across the twin networks due to the shared parameters. We used a minibatch size of 32 and Adam \cite{kingma:14} optimizer with the default parameters. For regularization, we observed that Batch Normalization \cite{sergey:14} provided better results than Dropout \cite{nitish:14}.

We used \textit{tanh} for GRU and \textit{ReLU} for all units as activation functions. We initialized all network weights from a normal distribution with variance $2.0/n$ \cite{he:15}, where $n$ is the number of units in previous layer. Although we use an adaptive optimizer, employing an exponential decay learning schedule helped improve performance. We trained EBM for a maximum of 150 epochs.

\subsection{Results}
\label{results}
We trained three baseline algorithms: Logistic Regression, LambdaMART \cite{burges:10}, and HypRank \cite{ruhi:18}, where Logistic Regression and LambdaMART were trained with the pairwise loss function, HypRank with the listwise loss function, and EnergyRank with both loss functions. For LambdaMART we used three different encoding schemes: one-hot vectors (OH), feature hashing (FH), and eigen-decomposition (ED). For HypRank we used $LSTM^{C}$, i.e, concatenating the hypothesis vectors and the BiLSTM output vectors as input to the feedforward layer since this was the best performing architecture.  

\begin{table}
\caption{Error-rates on labeled data both with and without information-state.}
\label{siri-errorrate} 
%
%
\begin{tabular}{p{3.5cm}P{3cm}P{2cm}P{3cm}P{2cm}}
\hline\noalign{\smallskip}
\bf Method & \bf Error Rate$^*$ & \bf p-value$^*$ & \bf Error Rate$^{**}$ & \bf p-value$^{**}$  \\
\hline\noalign{\smallskip}
$Logistic Regression$ & 41.1\% $\pm$ 0.5\%  & 0.7E-04 & 32.1\% $\pm$ 1.2\% & 1.2E-05 \\
$LambdaMART^{OH}$ & 36.5\% $\pm$ 0.3\% & 1.4E-05 & 22.3\% $\pm$ 0.1\% & 1.1E-05 \\
$EnergyRank_{list}^{EF}$ &  ---  & --- & 20.9\% $\pm$ 1.3\%  & 0.9E-05 \\
$LambdaMART^{FH}$ & 34.4\% $\pm$ 0.6\%  & 1.3E-05 & 20.2\% $\pm$ 0.1\%  & 1.1E-05 \\
$HypRank $ & 32.9\% $\pm$ 0.8\% & 1.6E-04 & 19.6\% $\pm$ 0.9\% & 2.3E-04 \\
$EnergyRank_{pair}^{HF}$ & --- & --- & 19.5\% $\pm$ 0.6\% & 1.6E-03 \\
$LambdaMART^{ED}$ & \textbf{29.7\% $\pm$ 0.3\%} & 0.9E-05 & 18.2\% $\pm$ 0.1\% & 1.2E-05 \\
$EnergyRank_{list}^{LF}$ & ---  & --- & 17.9\% $\pm$ 1.1\%  & 2.1E-03\\
$EnergyRank_{pair}^{LF}$ & ---  & --- & \textbf{17.5\% $\pm$ 0.8\%}  & 1.3E-05\\

\noalign{\smallskip}\hline\noalign{\smallskip}

\end{tabular}

$^*$ without information-state

$^{**}$ with information-state
\end{table}

\subsubsection{Error Rate}
\label{error-rate-sec}
We trained each model ten times with different seed and weight initializations, and we report the mean error rate. We use a two-sided T-test to compute p-value to establish statistical significance. Table-\ref{siri-errorrate} shows the results on the internal labeled dataset, with $\pm$ showing 95\% confidence intervals. We empirically show that information-state improves error-rates. EnergyRank results are not reported in experiments without information-state since it needs both understanding features and information-state to compute the affinity metric. The superscript of LambdaMART denotes the encoding scheme used. EnergyRank superscript denotes $\phi$ used: EF for Exponential Function, HF for Hinge Function, LF for Logistic Function, and subscript for pairwise/listwise loss paradigm.

\subsubsection{Relative Entropy}
We run the best performing methods: LambdaMART, HypRank, and EnergyRank models on two unlabeled datatsets, each of the size 80,000 sampled from different feature distributions. We use the score of the model's top predicted intent and group them into 21 buckets ranging from 0.0 to 1.0 with a step-size of 0.05. The raw counts obtained are normalized and interpolated to obtain a probability density function (PDF) of the scores. We measure the relative entropy to quantify the robustness of these algorithms to changes in feature distributions. The best performing EnergyRank model degrades in robustness when no affine-transform is applied ($EnergyRank_{pair}^{LF-NA}$) with a minimal drop in accuracy.

\begin{table}
\centering
\caption{Relative-Entropies on unlabeled data.}
\label{relentr} 
\begin{tabular}{p{3cm}P{3cm}}
\hline\noalign{\smallskip}
     \bf Method & \bf Relative Entropy \\ \hline
     $HypRank$ &  0.468 \\
     $EnergyRank_{pair}^{LF-NA}$ & 0.319\\
     $LambdaMART^{ED}$ &  0.168 \\
     $EnergyRank_{pair}^{LF}$ & \textbf{0.112} \\
\noalign{\smallskip}\hline\noalign{\smallskip}
\end{tabular}
\end{table}

\begin{figure*}
 \centering
   \includegraphics[scale=0.21]{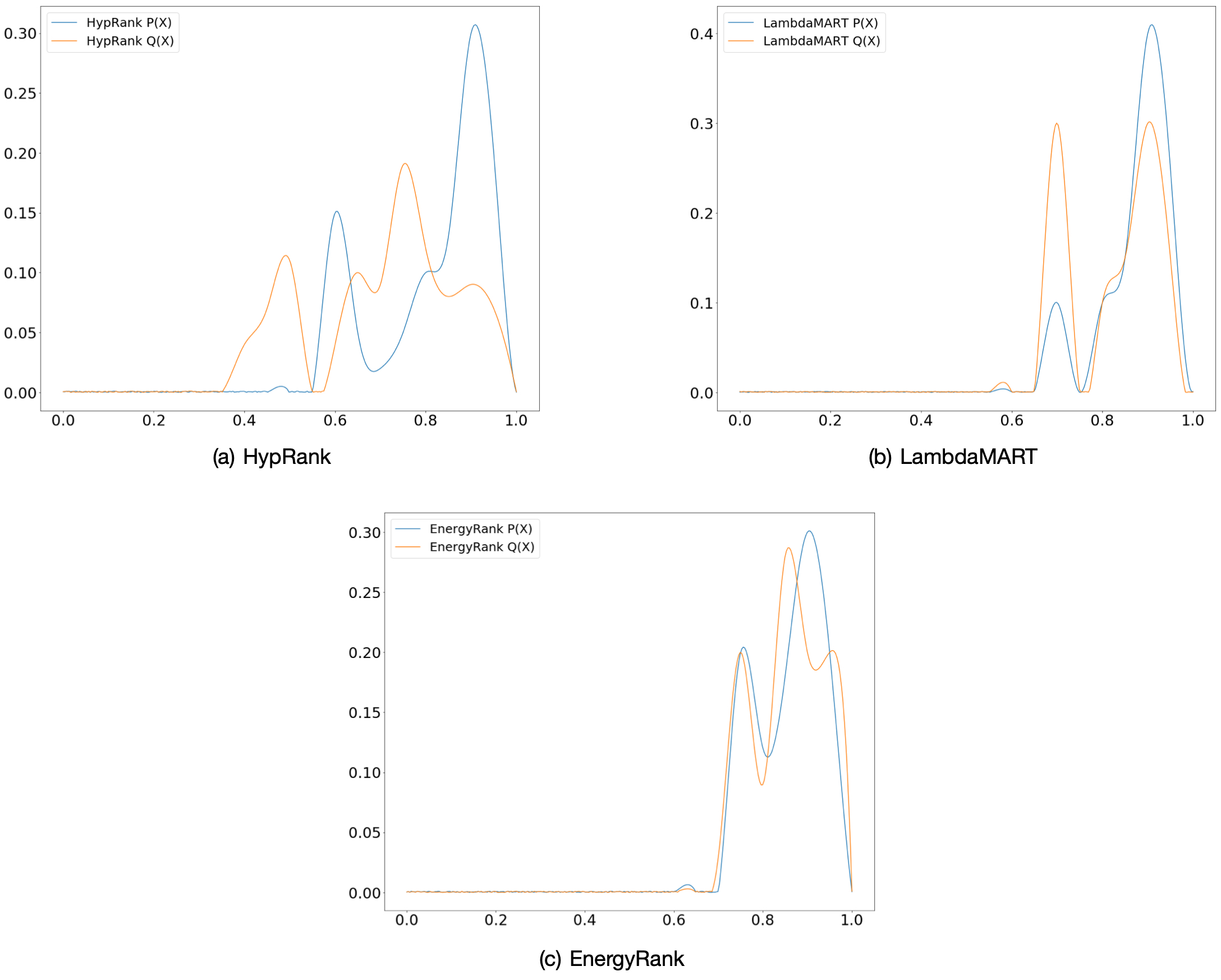}
   \caption{A visualization of the model's top intent score distributions as probability density function (PDF) corresponding to two different input distributions P(X) and Q(X).}
  \label{fig:rel-ent}
\end{figure*}

Figures \ref{fig:rel-ent}a, \ref{fig:rel-ent}b, and \ref{fig:rel-ent}c show the superimposition of the model's top intent output score PDFs of HypRank, LambdaMART, and EnergyRank respectively. The two output score PDFs in each superimposition correspond to P(X) and Q(X) input distributions. Table \ref{relentr} shows the relative-entropy which quantifies the difference between the two PDFs. EnergyRank with pairwise loss improves relative-entropy over LambdaMART with ED (best performing method among SOTAs, see Table \ref{siri-errorrate}) by 33.3\% and over HypRank by 76.1\%.

\section{Conclusion}
We have presented a novel ranking algorithm based on EBM for learning complex affinity metrics between extracted meaning from user requests and user information-state to choose the best response in a voice assistant. We described a Multisource DAE pretraining approach to obtain robust fused representations of data from different sources. We illustrated how our model is also capable of performing zero-shot decision making for predicting and selecting intents. We further evaluated our model against other SOTA methods for robustness and show our approach improves relative-entropy.

\section*{Acknowledgements}
We would like to thank our colleagues Barry Theobald, Arturo Argueta, Stephen Pulman, Steve Young, and Russ Webb for their insightful comments.

\nocite{*}
\bibliography{energy-rank}
\bibliographystyle{plain}

\end{document}